\def\year{2020}\relax
\documentclass[letterpaper]{article} 
\usepackage{aaai20}  
\usepackage{times}  
\usepackage{helvet} 
\usepackage{courier}  
\usepackage[hyphens]{url}  
\usepackage{graphicx} 
\usepackage{amsfonts}

\usepackage{multirow}
\usepackage{amsmath}
\usepackage{color}   
\usepackage{float}  

\usepackage{graphicx}  
\title{TANet: Robust 3D Object Detection from Point Clouds with Triple Attention} 

\author{ Zhe Liu$^{1}$, Xin Zhao$^{2}$, Tengteng Huang$^{1}$, Ruolan Hu$^{1}$, Yu Zhou$^{1}$,  Xiang Bai$^{1}$\thanks{Corresponding author} \\
$^1$Huazhong University of Science and Technology, Wuhan, China, 430074\\
$^2$Institute of Automation, Chinese Academy of Sciences, Beijing, China, 100190\\
m201772494@hust.edu.cn, xzhao@nlpr.ia.ac.cn, huangtengtng@hust.edu.cn,\\
huruolan@hust.edu.cn, yuzhou@hust.edu.cn, xbai@hust.edu.cn\\
}
 
\begin{document}

\maketitle

\begin{abstract}

In this paper, we focus on exploring the robustness of the  
3D object detection in point clouds, 
which has been rarely discussed in existing approaches. 
We observe two crucial phenomena: 
1) the detection accuracy of the hard objects, e.g., Pedestrians, is unsatisfactory, 2) when adding additional noise points, 
the performance of existing approaches decreases rapidly.
To alleviate these problems, 
a novel TANet is introduced in this paper, 
which mainly contains a Triple Attention (TA) module, 
and a Coarse-to-Fine Regression (CFR) module.
By considering the channel-wise, point-wise and voxel-wise attention jointly, 
the TA module enhances the crucial information of the target while suppresses the unstable cloud points.
Besides, the novel stacked TA further exploits the multi-level feature attention.
In addition, the CFR module boosts the accuracy of localization without excessive computation cost.
Experimental results on the validation set of KITTI dataset demonstrate that,
in the challenging noisy cases, 
i.e., adding additional random noisy points around each object,
the presented approach goes far beyond state-of-the-art approaches.
Furthermore, 
for the 3D object detection task of the KITTI benchmark, 
our approach ranks the first place on Pedestrian class, 
by using the point clouds as the only input. The running speed is around 29 frames per second.

\end{abstract}

\section{Introduction}

3D object detection in point clouds has a large number of applications in real scenes, 
especially for autonomous driving and augmented reality. 
On the one hand, point clouds provide reliable geometric structure information and precise depth, while how to utilize such information effectively is an essential issue. On the other hand, 
point clouds are usually unordered, sparse, and unevenly distributed, which poses great challenges for accurate object detection.

\begin{figure}[h]
\begin{center}
  \includegraphics[width=1.0\linewidth]{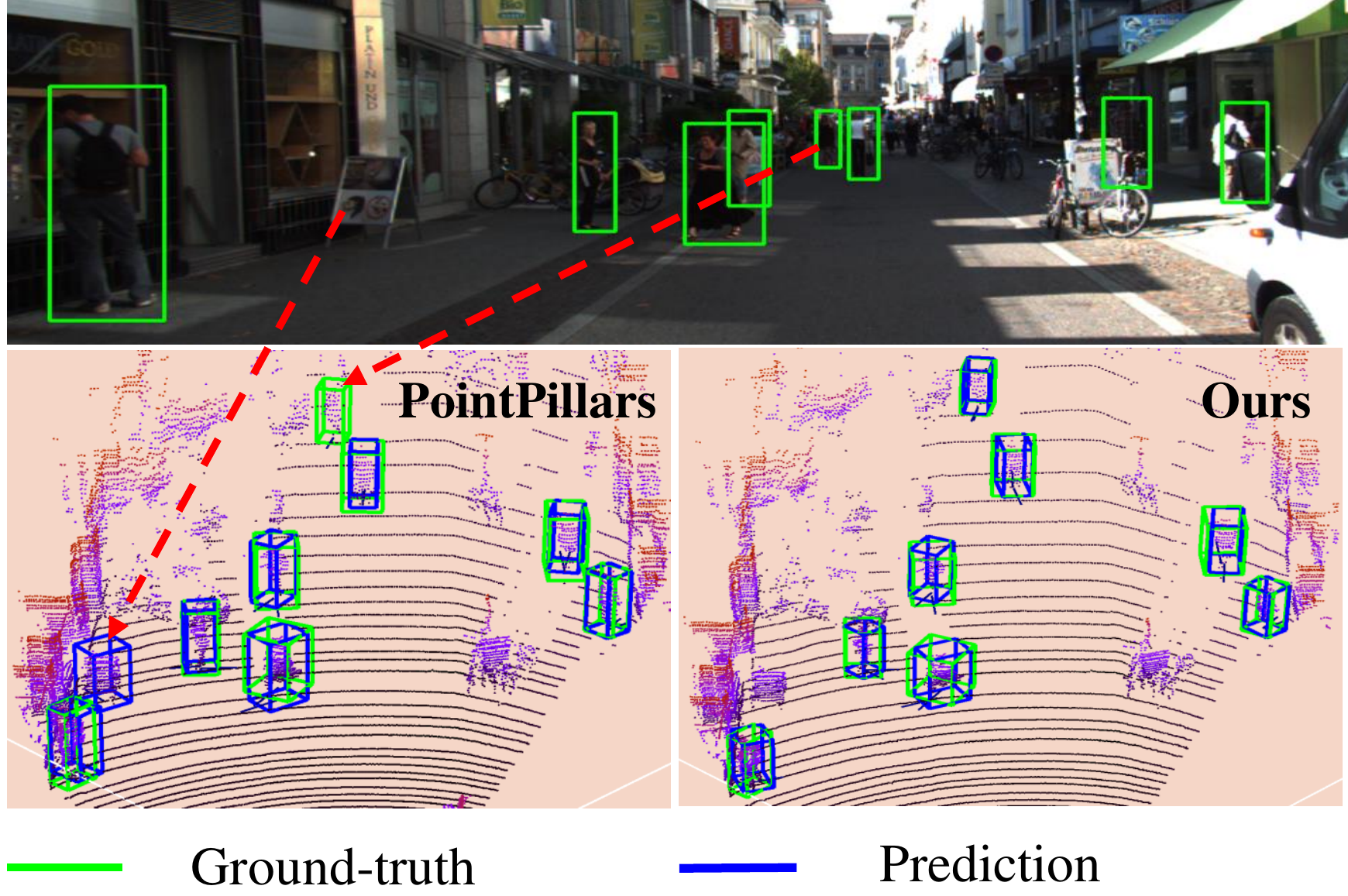}
\end{center}
\caption{Detection results for Pedestrians. The first row shows the corresponding 2D image. The second row demonstrates the 3D detection results produced by PointPillars and our method, respectively. We highlight the missed and false detection in PointPillars with red arrows. }
\label{Intro}
\end{figure}

In recent years, several approaches based on point clouds have been proposed in the 3D object detection community. PointRCNN~\cite{shi2019pointrcnn} directly operates on the raw point clouds, extracts the features by PointNets~\cite{qi2017pointnet,qi2017pointnet++} and then estimates final results by two-stage detection networks. 
VoxelNet~\cite{zhou2018voxelnet}, SECOND~\cite{yan2018second} and PointPillars~\cite{lang2019pointpillars} convert the point clouds to the regular voxel grid and apply a series of convolutional operations to 3D object detection.

\begin{figure*}[t]
\begin{center}
  \includegraphics[width=0.9\linewidth]{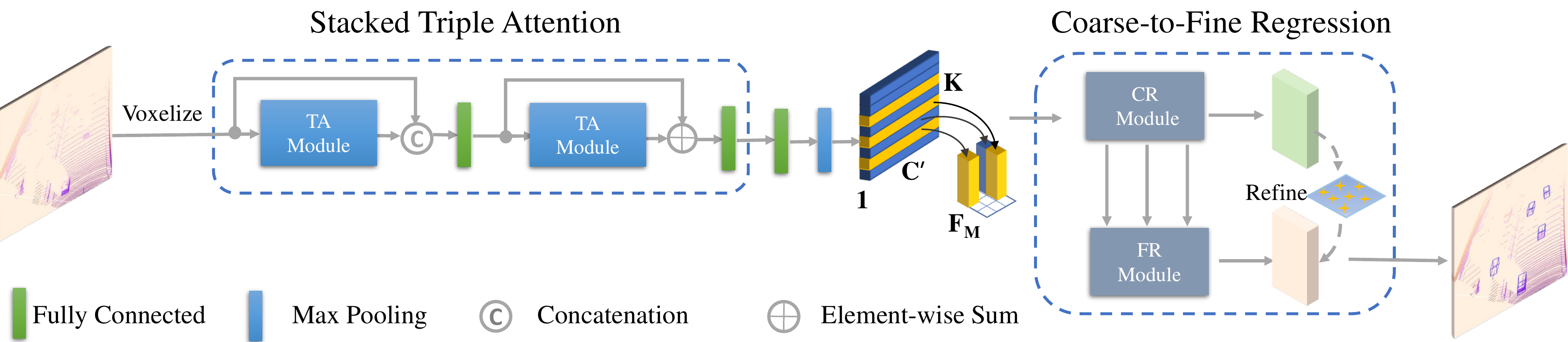}
\end{center}
\caption{The full pipeline of TANet. 
First, we equally divide the point clouds into a voxel grid consisting of a set of voxels. Then, the stacked triple attention separately process each voxel to obtain a more discriminative representation. Subsequently, a compact feature representation for each voxel is extracted by aggregating the points inside it in a max-pooling manner. We arrange the voxel feature according to its original spatial position in the grid and thus lead to a feature representation for the voxel grid in the shape of $C' \times H \times W$. Finally, the coarse-to-fine regression is  employed to generate the final 3D bounding boxes.}

\label{Architecture}
\end{figure*}

Although existing approaches have reported promising detection accuracy,  the performance is still unsatisfactory in challenging cases, especially hard objects such as pedestrians.
As shown in Fig.~\ref{Intro}, PointPillars~\cite{lang2018pointpillars} misses a pedestrian and provides a false positive object. We reveal the intrinsic reasons from two aspects, 
1) pedestrians have a smaller scale than cars, which makes fewer valid points scanned on them through LiDAR.
2)  pedestrians frequently appear in a variety of scenes, so various background objects, such as trees, bushes, poles, \textit{etc.}, might be close to the pedestrians, 
which results in enormous difficulty to recognize them correctly. Hence, detecting objects in the complex point clouds is still an extremely challenging task.

In this paper, we present a novel architecture named Triple Attention Network~(TANet), as shown in Fig~\ref{Architecture}. 
The straightforward motivation is that, in the cases of severe noises, a set of informative points can supply sufficient cues for the subsequent regression. In order to capture such informative cues, a TA module is introduced to enhance the discriminative points and suppress the unstable points. Specifically, the point-wise attention, channel-wise attention are learned respectively, and they are combined by element-wise multiplication. Besides, we also consider the voxel-wise attention, which represents the global attention of a voxel.

In the noisy cases, only applying a single regressor, e.g., one stage RPN, for 3D bounding box localization is unsatisfactory. To address this issue, we introduce an end-to-end trainable coarse-to-fine regression (CFR) mechanism. The coarse step provides a rough estimation of the object following~\cite{zhou2018voxelnet,lang2019pointpillars}. 
Then we present a novel Pyramid Sampling Aggregation (PSA) fusion approach, which supplies cross-layer feature maps. The refinement is implemented upon the fused cross-layer feature map to obtain the finer estimation.

Both the TA module and CFR mechanism are crucial for the robustness of 3D detectors, which is very important for the real scenario of automatic driving. Since not all the data in KITTI dataset is troubled by noises,
in the experimental evaluation, we simulate the noisy cases by adding random points around each object. Extensive experiments demonstrate that our approach greatly outperforms the state-of-the-art approaches. Besides, our approach achieves the best results on the Pedestrian class of the KITTI benchmark, which further verifies the robustness of the presented detector.

In summary, the key contributions of the proposed method lie in:

1) We introduce a novel Triple Attention module, which takes the channel-wise, point-wise, and voxel-wise attention into consideration jointly, then the stacked operation is performed to obtain the multi-level feature attention, 
and hence the discriminative representation of the object is acquired. 

2) We propose a novel coarse-to-fine regression, based on the coarse regression results, the fine regression is performed on the informative fused cross-layer feature map. 

3) Our approach achieves convincing experimental results in the challenging noisy cases, and the quantitative comparisons on the KITTI benchmark illustrate that our method achieves state-of-the-art performance and promising inference speed.

\section{Related Work}

With the rapid development of computer vision, much effort has been devoted to detect 3D objects from multi-view images, 
which can be roughly classified into two categories: 
the front view based approaches \cite{song2015joint,chen2016monocular,mousavian20173d},
and the bird's eye view based approaches \cite{chen2017multi,ku2018joint,yang2018pixor,simon2018complex,li2016vehicle,yang2018hdnet}. However, it is difficult for these methods to localize the objects accurately due to the loss of depth information. 

Recently, the research trend has gradually shifted from the RGB image to point cloud data. Detecting 3D objects based on the voxel grid has been widely concerned
\cite{engelcke2017vote3deep,wang2015voting}. 
In these approaches, 3D point cloud space is divided into voxels equally, 
and only the non-empty voxels are encoded for computational efficiency. VoxelNet \cite{zhou2018voxelnet}, SECOND \cite{yan2018second} and PointPillars \cite{lang2019pointpillars} convert point clouds to a regular voxel grid and learn the  representation of each voxel with the Voxel Feature Encoding (VFE) layer.
Then, the 3D bounding boxes are computed by a region proposal network based on the learned voxel representation. In contrast to PointPillars, we focus on leveraging channel-wise, point-wise, and voxel-wise attention of point clouds to learn a more discriminative and robust representation for each voxel. To our best knowledge, 
the proposed method is the first one to design the attention mechanism suitable for the 3D object detection task. In addition, our method utilizes stacked Triple Attention~(TA) modules to capture the multi-level feature attention.

In the 3D object detection task, 
the voxel grids based approaches, e.g. Voxelnet, PointPillars and SECOND, frequently adopt one-stage detection network~\cite{liu2016ssd,fu2017dssd,shen2017dsod}, which can process more than 20 frames per second. In contrast, the raw point clouds based approaches, e.g., PointRCNN \cite{shi2019pointrcnn}, utilize two-stage architecture~\cite{girshick2015fast,ren2015faster,lin2017feature,he2017mask}. 
Although these methods lead to better detection performance, 
they run at relatively slower speeds with less than 15 frames per second.
Motivated by RefineDet \cite{zhang2018single},
we propose an end-to-end trainable coarse-to-fine regression mechanism. Our goal is to seek detection methods that can achieve a better trade-off better accuracy and efficiency.

\section{3D Object Detection with TANet}
\label{Our_methods}

In this section, we introduce TANet for 3D object detection based on voxels, which is an end-to-end trainable neural network. As shown in Fig.~\ref{Architecture}, our network architecture consists of two main parts: the Stacked Triple Attention and the Coarse-to-Fine Regression. 

Before introducing the technical details, we present several basic definitions of 3D object detection. A point set in 3D space is defined as $\mathbf{P}=\{\mathbf{p}_i=[x_i, y_i, z_i, r_i]^T \in {\mathbb{R}} \}_{i=1,2,...,M}$, where $x_i, y_i, z_i$ denote the coordinate values of each point along the axes X, Y, Z, 
$r_i$ is the laser reflection intensity that can be treated as an extra feature, 
and $M$ is the total number of points. Given an object in 3D space, it is represented by a 3D bounding box $(c_x, c_y, c_z, h, w, l, \theta )$, including its center $c_x, c_y, c_z$, size $h, w, l$, and orientation $\theta$ that indicates the heading angle around the up-axis.

\subsection{Stacked Triple Attention}

We suppose that a point cloud $\mathbf{P}$ in the 3D space has the range of $W^*, H^*, D^*$ along the X, Y, Z axes respectively. $\mathbf{P}$ is equally divided into a specific voxel grid. Each voxel has the size of $v_W^*, v_H^*$, $v_D^*$. Consequently, the voxel grid is of size $W= W^*/v_W^*, H = H^*/v_H^*$, $D= D^*/v_D^*$. Note that $D$ is always 1, since the voxel grid is not discretized along Z axis in the implementation.

As the point clouds are often sparse and unevenly distributed, a voxel has a variable number of points. 
Let $N$ and $C$ denote the maximum number of points of each voxel and the channel number of each point feature, respectively. A voxel grid $\mathbf{V}$ that consists of $K$ voxels can be defined as
$\mathbf{V}=\{V^1,...,V^K\}$, where $V^k \in {\mathbb{R}}^{N\times C}$ indicates the $k$-th voxel of $\mathbf{V}$. 

\begin{figure}[t]
\begin{center}
  \includegraphics[width=0.9\linewidth]{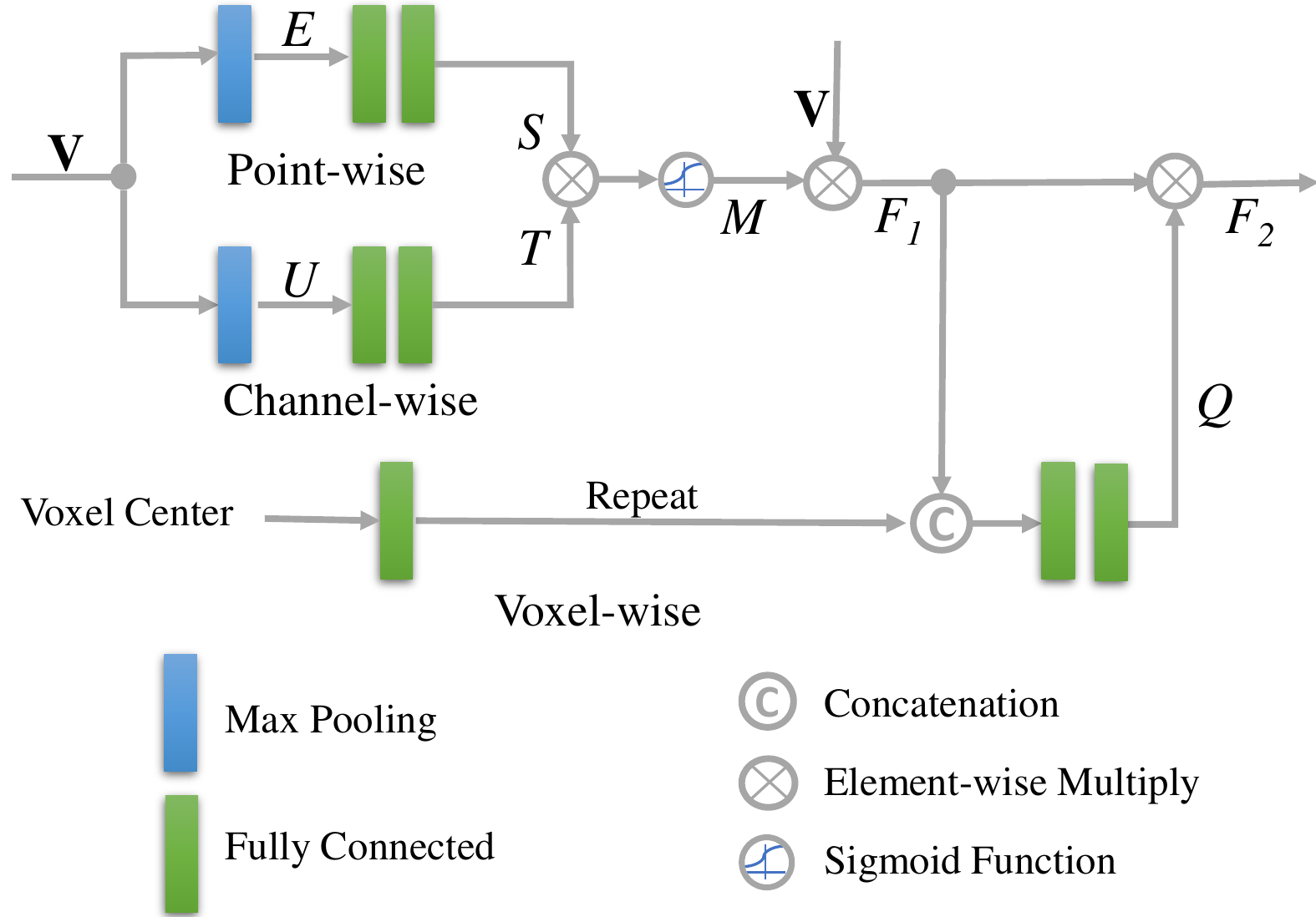}
\end{center}
\caption{The architecture of TA module. }
\label{Voxel_feature}
\end{figure}

\textbf{Point-wise Attention.}  
Given a voxel $V^k$, we perform max-pooling to aggregate point features across the channel-wise dimensions, resulting in its point-wise responses $E^{k} \in {\mathbb{R}}^{N \times 1}$. To explore the spatial correlation of points, following the \emph{Excitation} operation~\cite{hu2018squeeze}, 
two fully-connected layers are employed to encode the global responses, i.e.: 
\begin{equation}
S^k = W_2 \delta(W_1 E^k),
\end{equation}
where $W_1 \in {\mathbb{R}}^{r \times N}$, $W_2 \in {\mathbb{R}}^{N \times r} $ are the weight parameters of two fully-connected layers, respectively. $\delta$ is the ReLU activation function.
$S^k \in {\mathbb{R}}^{N \times 1}$ is the point-wise attention of $V^k$.
As shown in Fig.~\ref{Voxel_feature},
the upper branch of the attention module is designed
for describing the spatial correlation among the points inside each voxel.

\textbf{Channel-wise Attention.} 
Similar to the strategy for estimating the point-wise attention for a voxel, we compute the channel-wise attention with the middle branch of the attention module as shown in Fig.~\ref{Voxel_feature}.
A max-pooling operation is performed to aggregate the channel features across their point-wise dimensions, which obtains the channel-wise responses of a voxel $V^k$ to be $U^k \in {\mathbb{R}}^{1 \times C}$.
Then, we compute $T^k = W'_2 \delta(W'_1 {(U^k)}^T)$.
$W'_1 \in {\mathbb{R}}^{r \times C}$ and $ W'_2 \in {\mathbb{R}}^{C \times r} $ represent the importance of feature channels for each voxel. 

Given the $k$-th voxel $V^k$, we obtain the attention matrix ${M}^k\in {\mathbb{R}}^{N\times C}$ that combines spatial-wise attention ${S}^k$ and channel-wise attention ${T}^k$ through the element-wise multiply, i.e. : 
\begin{equation}
{M}^k= \sigma({S}^k \times {T}^k)
\end{equation}
where $\sigma$ denotes the \emph{sigmoid} function, 
which is employed to normalize the values of the attention matrix to the range of $[0,1]$.
Thus, a feature representation ${F}^k_1= {M}^k \odot {V}^k \in {\mathbb{R}}^{N\times C}$ can be obtained,
which  properly weights the importance of all the points inside a voxel across the point-wise and channel-wise dimensions. 


\textbf{Voxel-wise Attention.} 
The voxel-wise attention is further employed to judge the importance of the voxels.
We first average the coordinates of all points inside each voxel as the voxel center, which can provide accurate location information. 
Then the voxel center is transformed into a higher dimension through a fully-connected layer, 
and it is combined with ${F}^k_1$ in a concatenation fashion. 
Voxel-wise attention weight is defined as ${Q}=[q^1,...,q^k,...,q^K] \in \mathbb{R}^{K \times 1 \times 1}$, 
where $q^k$ is obtained by compressing the point-wise and channel-wise dimensions to 1 via two fully-connected layers, respectively. 
Finally, a more robust and  discriminative voxel feature is obtained by ${F}^k_2= q^k \cdot {F}^k_1$.

Through all the above operations, 
the feature representation ${F}^k_2$ enhances the crucial features,
which contributes significantly to our tasks while suppresses the irrelevant and noising features. 
For simplicity, we name the module integrating these three types of attention as Triple Attention~(TA).

\textbf{Stacked TA.}
As shown in Fig.~\ref{Architecture},
in our approach, 
two TA modules are stacked to exploit the multi-level feature attention. 
The first one directly operates on the original features of the point clouds, while the second one works on the higher dimensional features. 
For each TA module, we concatenate/sum its output with its input to fuse more feature information.  
Then the higher-dimensional feature representation is obtained via a fully-connected layer. 
Finally, a max-pooling operation is used to aggregate all the point features of each voxel, which is treated the input of the CFR.


\subsection{Coarse-to-Fine Regression}

We employ a Coarse Regression (CR) module and a Fine Regression (FR) module for 3D box estimation. The details of these two modules are presented in the following.

\begin{figure}[t]
\begin{center}
\includegraphics[width=1.0\linewidth]{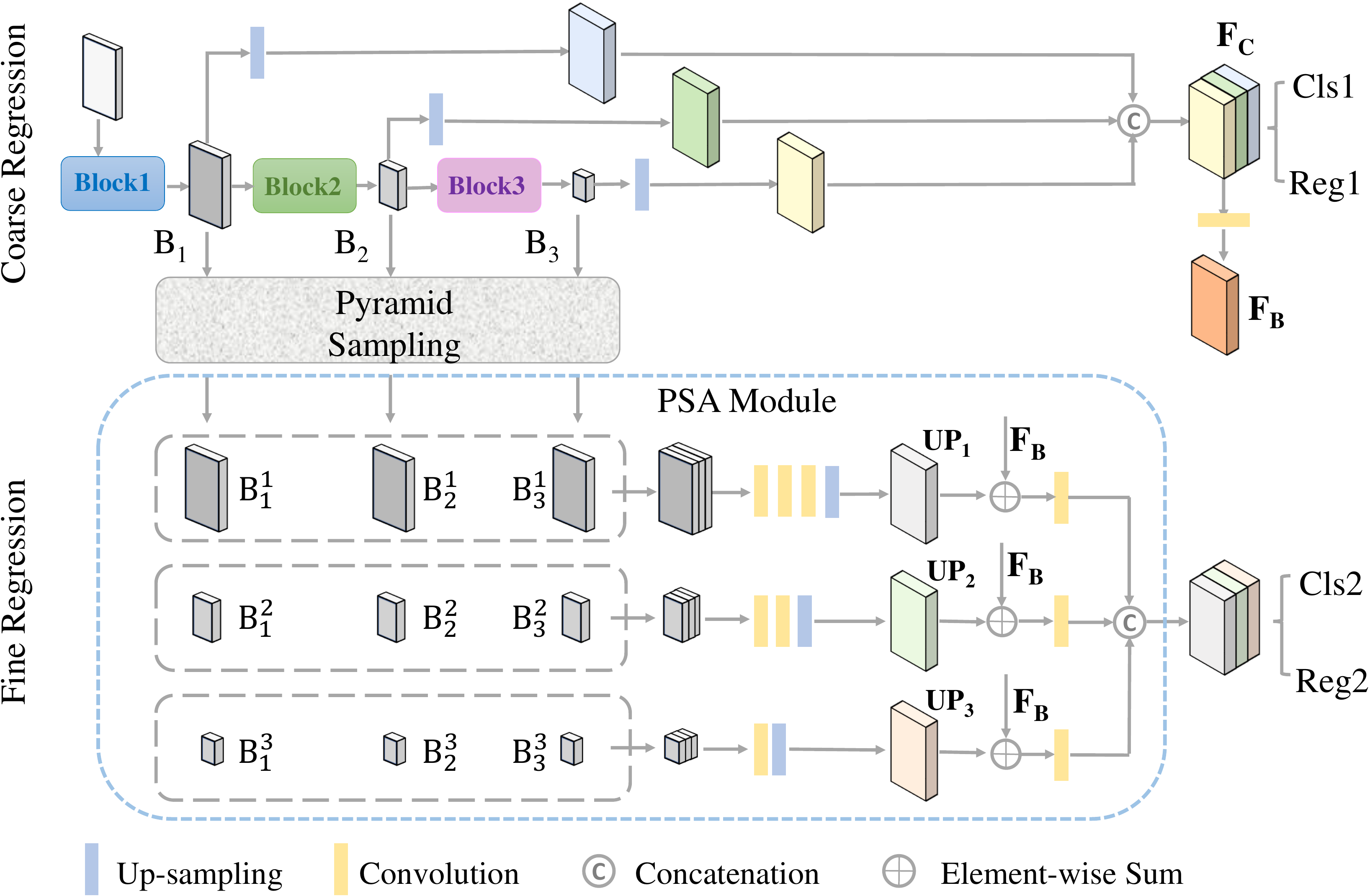}
\end{center}
\caption{The architecture of Coarse-to-Fine Regression. The Pyramid Sampling  indicates a  series of  down-sampling and up-sampling operations, which can be achieved via  the pooling  and the transposition convolution. 
}
\label{PSA}
\end{figure}

The CR module uses a similar architecture with \cite{zhou2018voxelnet,yan2018second,lang2018pointpillars}. To be specific, 
as shown in the top module of Fig.~\ref{PSA}, 
the output of Block1, Block2 and Block3 is denoted as $B_{1}$, $B_{2}$ and ${B}_{3}$,  
and the shapes of these blocks are $(C', H/2, W/2)$, $(2C', H/4, W/4)$ and $(4C', H/8, W/8)$, respectively.
The CR module generates a feature map $\mathbf{F_C}$ with the size of $(6C', H/2, W/2)$, then follows the classification and regression branches, which provides the coarse boxes for FR module.

Based on the output of the CR module, 
the Pyramid Sampling Aggregation (PSA) module is leveraged to provide cross-layer feature maps, which is shown in the bottom module of  Fig.~\ref{PSA}. The high-level features supply larger receptive fields and richer semantic information, while the low-level features have a larger resolution. Thus, the cross-layer feature maps effectively capture multi-level information, leading to a more comprehensive and robust feature representation for objects. Specifically, a feature pyramid $\{{B}_{1}^{1}, {B}_{1}^{2}, {B}_{1}^{3}\}$ is achieved based upon ${B}_{1}$,  
where ${B}_{1}^{1}$ is equivalent to ${B}_{1}$.
${B}_{1}^{2}$ and ${B}_{1}^{3}$ 
are obtained by two down-sampling operations  performing on ${B}_{1}$, respectively.
${B}_{1}^{2}$ has the same size as ${B}_{2}$, 
and ${B}_{1}^{3}$ has the same size as ${B}_{3}$. Similarly, an up-sampling and a down-sampling are operated on ${B}_{2}$ to obtain $\{{B}_{2}^{1}, {B}_{2}^{2}, {B}_{2}^{3}\}$. In addition, $\{ B_{3}^{1}, {B}_{3}^{2}, {B}_{3}^{3}\}$ is obtained by two up-sampling operations based on $B_{3}$.

To make full use of the cross-layer features, 
we concatenate ${B}_{1}^{i}$, ${B}_{2}^{i}$ and ${B}_{3}^{i}$ for $i=1,2,3$, respectively. Then a series of convolution operations followed by an up-sampling layer are executed, which results in the feature maps $UP = \{UP_1, {UP}_2, {UP}_3\}$, they have the same shape of $(2C' \times H/2 \times W/2)$. 

In addition, the enriched feature maps from the PSA module are combined with the semantic information from CR module.
Specifically,
a $1 \times 1$ convolution is employed to transform $F_C$ into $F_B$,  
$F_B$ has the same dimension with feature maps in $UP$.
Then each feature map in $UP$ is combined with $F_B$ by element-wise addition. 
A $3 \times 3$ convolution layer is performed on each fused feature map. 
$F_R$ is obtained by concatenating the modulated features in ${UP}$, 
which serves as the feature map for Fine Regression.
The regression branch of FR regards the coarse boxes of CR as the new anchors to regress the 3D bounding box, and perform the classification.

\begin{table*}[h]
\scriptsize
\centering
\begin{tabular}{l|c|c|c|c|c|c|c|c|c|c|c|c|c}
\hline
\multirow{2}{*}{Method} 
& \multicolumn{1}{c|}{\multirow{2}{*}{\shortstack{Number of\\noise points}}} 
& \multicolumn{4}{c|}{Cars} 
& \multicolumn{4}{c|}{Pedestrians}
& \multicolumn{4}{c}{Cyclists}  \\

\cline{3-14}
& \multicolumn{1}{c|}{} & Easy & Moderate & Hard & 3D mAP & Easy & Moderate & Hard & 3D mAP & Easy & Moderate & Hard & 3D mAP  \\
\hline
\hline
PointRCNN~\shortcite{shi2019pointrcnn}  &0 &88.26	&77.73	&76.67 &\textbf{80.89}
 &65.62	&58.57	&51.48	&58.56 &82.76 &62.83 &59.62 &68.40 \\
PointPillars~\shortcite{lang2019pointpillars}  &0 &87.50	&77.01 &74.77 &79.76 &66.73 &61.06 &56.50 &61.43 &83.65	&63.40 &59.71 &68.92 \\
Ours  &0 &88.21	&77.85 &75.62 &80.56  &70.8	&63.45 &58.22 &\textbf{64.16} &85.98	&64.95 &60.40 &\textbf{70.44} \\
\hline
\hline
PointRCNN~\shortcite{shi2019pointrcnn}  &20  &88.24	&76.95 &74.73 &79.97 &62.00	&56.17 &49.52 &55.90 &81.55 &61.98	&57.20 &66.91 \\
PointPillars~\shortcite{lang2019pointpillars}  &20 &87.21 &76.74 &74.54 &79.50 &64.44 &59.02 &55.00 &59.49 &82.66 &62.52 &58.23 &67.80 \\
Ours  &20 &88.17 &77.68 &75.31 &\textbf{80.39} &69.98 &62.70 &57.65 &\textbf{63.44} &85.55 &64.06 &60.03 &\textbf{69.88} \\
\hline
PointRCNN~\shortcite{shi2019pointrcnn}  &50  &87.99	&76.66 &74.16 &79.60 &58.12	&51.23 &45.30	&51.55 &79.49	&60.63	&56.3 &65.47\\
PointPillars~\shortcite{lang2019pointpillars}  &50 &87.07	&76.60	&69.05	&77.57
 &62.75	&57.32 &52.25 &57.44  &81.98 &61.15 &56.53 &66.55 \\
Ours  &50 &87.97&77.29 &74.4 &\textbf{79.89}  &69.37	&62.5 &56.54 &\textbf{62.80} &84.85	&63.54 &58.48 &\textbf{68.96} \\
\hline
PointRCNN~\shortcite{shi2019pointrcnn}  &100  &87.56 &75.98	&69.34 &77.63 &55.57 &48.35 &42.88&	48.93 &76.77 &56.66	&52.92 &62.12 \\
PointPillars~\shortcite{lang2019pointpillars}  &100 &86.62 &76.06 &68.91 &77.20 &60.31 &55.17 &49.65 &55.04  &80.97	&58.02 &54.6 &64.53 \\
Ours  &100 &87.52 &76.64 &73.86 &\textbf{79.34}  &67.30 &60.77 &54.45 &\textbf{60.84}  &84.53 &61.64 &57.44 &\textbf{67.87}\\
\hline
\end{tabular}
\caption{Performance comparison with PointRCNN and PointPillars for 3D object detection task on the KITTI validation set for \textit{Cars}, \textit{Pedestrians} and \textit{Cyclists}. 3D mAP represents the mean average precision of each category.}
\label{Noise_Experiments}
\end{table*}

\subsection{Loss Function}

The multi-task loss function is employed for jointly optimizing the CR module and the FR module. 
The offsets of the bounding box regression between a prior anchor $a$ and the ground-truth box $g$ can be computed as:
\begin{equation}
\begin{aligned}
\Delta_x^g &= \frac{x_g-x_a}{d_a}, \Delta_y^g = \frac{y_g-y_a}{d_a}, \Delta_z^g = \frac{z_g-z_a}{h_a} \\
\Delta_w^g &= log(\frac{w_g}{w_a}), \Delta_l^g = log(\frac{l_g}{l_a}), \Delta_h^g = log(\frac{h_g}{h_a})  \\
\Delta_{\theta}^g&=\theta_g-\theta_a,
\end{aligned}
\end{equation}
where $d_a =\sqrt{(w_a)^2+(l_a)^2}$. For simplicity, the residual vector  $\mathbf{\Delta_g}=(\Delta_x^g, \Delta_y^g, \Delta_z^g, \Delta_w^g, \Delta_l^g, \Delta_h^g, \Delta_{\theta}^g)$  is defined as the regression ground-truth. Similarly,
$\mathbf{\Delta_p}=(\Delta_x^p, \Delta_y^p, \Delta_z^p, \Delta_w^p, \Delta_l^p, \Delta_h^p, \Delta_{\theta}^p)$ indicates the offsets between the prior anchors and the predicted boxes.  SmoothL1~\cite{girshick2015fast} is used as our 3D bounding box regression loss $L_{reg}$. Besides, angle loss is employed for better restricting the orientation of 3D bounding box. Note that when the orientation angle of a 3D object is shifted $\pm\pi$ radians, it does not change the estimation of localization. Hence, the \emph{sine} function is introduced to encode the loss of the orientation angle $\theta$ following~\cite{yan2018second}. Considering that the number of positive samples and negative samples is imbalanced, the Focal Loss~\cite{lin2017focal} is adopted as the classification loss $L_{cls}$. The superscript $\mathcal{C}$ and $\mathcal{R}$ represent the CR module and the FR module, respectively. It should be noted that the FR module leverages the coarse bounding boxes as the new anchor boxes, which is different from
the  CR module that utilizes the prior anchor boxes. The total loss function can be defined as:
\begin{equation}
\begin{aligned}
L_{total}&= \alpha L_{cls}^{\mathcal{C}} + \beta \frac{1}{N_{pos}^{\mathcal{C}}} \sum L_{reg}^{\mathcal{C}}(\mathbf{\Delta_p^\mathcal{C}},\mathbf{\Delta_g^\mathcal{C}}) + \\
&\lambda\{\alpha L_{cls}^{\mathcal{R}} + \beta \frac{1}{N_{pos}^\mathcal{R}} \sum L_{reg}^{\mathcal{R}}(\mathbf{\Delta_p^\mathcal{R}},\mathbf{\Delta_g^\mathcal{R}}) \}
\end{aligned}
\end{equation}
where $N_{pos}^{\mathcal{C}}$ and $N_{pos}^{\mathcal{R}}$ stand for the numbers of positive anchors in the CR module and FR module, respectively. $\alpha$ and $\beta$ represent the balance weights for the classification loss and the regression loss, respectively. $\lambda$ is used to balance the weight for the CR module and FR module.

\section{Experiments}

\begin{table*}[h]
\scriptsize
\centering
\begin{tabular}{l|c|c|c|c|c|c|c|c|c|c|c}
\hline
\multirow{2}{*}{Method} 
&\multicolumn{3}{c|}{Cars} 
& \multicolumn{3}{c|}{Pedestrians} 
& \multicolumn{3}{c|}{Cyclists} 
& \multicolumn{1}{c|}{\multirow{2}{*}{Modality}}
& \multicolumn{1}{c}{\multirow{2}{*}{3D mAP~(\%)}} \\
\cline{2-10}
& Easy & Moderate & Hard & Easy & Moderate & Hard & Easy & Moderate & Hard & \multicolumn{1}{c|}{}  \\
\hline
\hline
MV3D~\shortcite{chen2017multi}  &71.09  &62.35  &55.12 &- &- &- &- &- &- &LiDAR \& Image   &- \\
UberATG-ContFuse~\shortcite{liang2018deep} &82.54 &66.22 &64.04 &- &- &- &- &- &- &LiDAR \& Image  &-  \\
PC-CNN-V2~\shortcite{8461232}   &84.33 &73.80  &64.83 &- &- &- &- &- &- &LiDAR {\&} Image  &- \\
AVOD~\shortcite{ku2018joint} &73.59 &65.78  &58.38 &38.28 &31.51 &26.98 &60.11 &44.90 &38.80 &LiDAR \& Image  &48.70   \\
AVOD-FPN~\shortcite{ku2018joint} &81.94 &71.88 &66.38 &50.80 &42.81 &40.88 &64.00 &52.18 &46.61 &LiDAR \& Image   &57.50  \\
F-Pointnet~\shortcite{qi2018frustum} &81.20 &70.39 &62.19 &51.21 &44.89 &40.23 &71.96 &56.77 & 50.39 &LiDAR \& Image  &58.80   \\
\hline
MV3D~(LiDAR)~\shortcite{chen2017multi}  &66.77 &52.73 &51.31 &- &- &- &- &- &- &LiDAR &- \\
Voxelnet~\shortcite{zhou2018voxelnet} &77.47 &65.11 &57.73 &39.48 &33.69 &31.51 &61.22 &48.36 &44.37 &LiDAR &50.99 \\
SECOND~\shortcite{yan2018second} &83.13 &73.66 &66.20 &51.07 &42.56 &37.29 &70.51 &53.85 &46.90 &LiDAR &58.35  \\
PointPillars~\shortcite{lang2019pointpillars}  &79.05 &74.99 &68.30 &52.08 &43.53&41.49&\textbf{75.78} &59.07&52.92  &LiDAR &60.80 \\
PointRCNN~\shortcite{shi2019pointrcnn}  &\textbf{85.94} &\textbf{75.76} &\textbf{68.32}  &49.43 &41.78 &38.63 & 73.93 &59.60 &\textbf{53.59}  &LiDAR &60.78 \\
\hline
\hline
TANet~(Ours) &83.81 &75.38 &67.66 &\textbf{54.92} &	\textbf{46.67} &\textbf{42.42}  &73.84 &\textbf{59.86} &53.46  &LiDAR  &\textbf{62.00}  \\
\hline
\end{tabular}
\caption{Performance comparison with previous approaches for 3D object detection task on the KITTI test split for \textit{Cars}, \textit{Pedestrians} and \textit{Cyclists}. 3D mAP represents the mean average precision of all three categories on 3D object detection.}
\label{3D_Object_Detection}
\end{table*}

\begin{figure*}[t]
\begin{center}
  \includegraphics[width=0.96\linewidth]{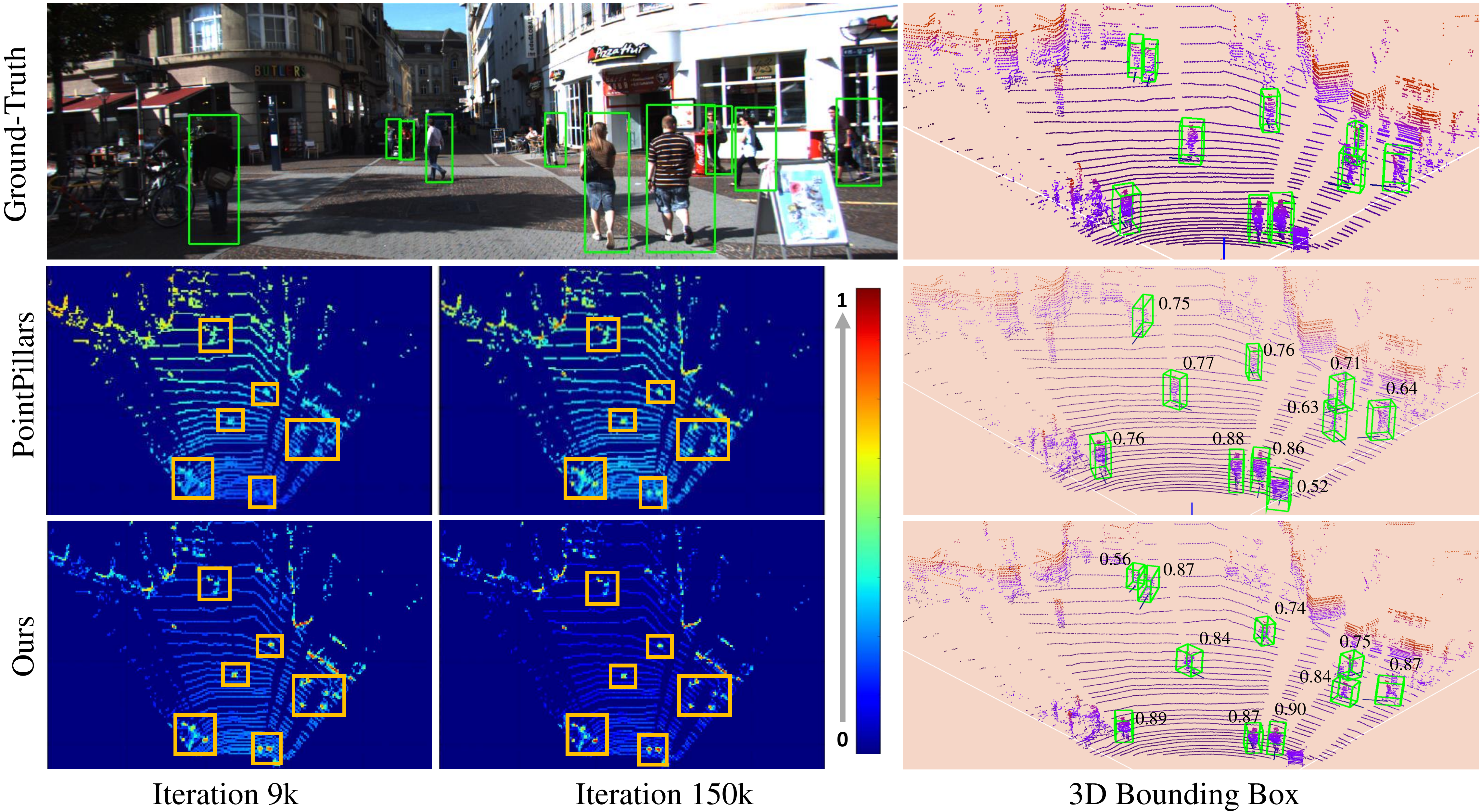}
\end{center}
\caption{Visualization of learned feature map and predicted confidence score. The first row shows the ground-truth detection on 2D image and 3D point clouds. The second and third row illustrates the learned feature map and the predicted confidence score of PointPillars and our methods, respectively. We highlight some crucial areas for each feature map with a yellow rectangular box.}
\label{Vis_Attention}
\end{figure*}

\subsection{Experimental Dataset and Evaluation Metrics}
\label{experiment4_1}

All of the experiments are conducted on the KITTI dataset~\cite{geigerwe}, which contains 7481 training samples and 7518 test samples. Since the access to the ground truth for the test set is not available, we follow~\cite{qi2018frustum,chen2017multi} to split the training samples into a training set consisting of 3712 samples and a validation set consisting of 3769 samples. When evaluating the performance on the test set, the training samples are re-split into the training and validation set according to the ratio of around 5:1. Our results are reported on both the KITTI validation set and test set for \textit{Cars}, \textit{Pedestrians} and \textit{Cyclists} categories. For each category, three difficulty levels are involved (\textit{Easy}, \textit{Moderate} and \textit{Hard}), which depend on the size, occlusion level and truncation of 3D objects.

The mean Average Precision (mAP) is utilized as our evaluation metric. For fair comparisons, we adopt the official evaluation protocol. Concretely, the IoU threshold is set to 0.7 for \textit{Cars} and the IoU threshold to 0.5 for \textit{Pedestrians} and \textit{Cyclists}.

\subsection{Implementation Details}
\label{experiment4_2}
For data augmentation, the points inside a ground-truth 3D bounding box along the Z-axis are rotated following the uniform distribution of [-$\pi$/4, $\pi$/4] for orientation varieties. Besides, we further randomly flip the point clouds in 3D boxes along the X-axis. Random scaling in the range of [0.95, 1.05] is also applied.  Ground-truth boxes are randomly sampled and placed into raw samples to simulate the scenes crowded with multiple objects~\cite{yan2018second}. 

Adaptive Moment Estimation~(Adam)~\cite{kingma2014adam} is used for optimization with the learning rate of 0.0002. And our model is trained for about 200 epochs with a mini-batch size of 2. In all of our experiments, we randomly sample fixed $N$ points for voxels containing more than $N$ points. For the voxels containing less than $N$ points, we simply pad them with zeros. In our settings, a large value of $N$ is selected to be 100 for capturing sufficient cues to explore the spatial relationships. The dimension of the feature map for each voxel $\mathbf{F_M}$ is 64 (e.g., $C'=64$). All of our experiments are evaluated on a single Titan  V GPU card. In our experiments, we set $\alpha$, $\beta$, and $\lambda$ to 1.0, 2.0 and 2.0 for total loss, respectively. For more implementation details, please refer to our supplemental material.

\subsection{Evaluation on KITTI dataset}
\label{experiment4_3}
In this part, our method is compared with the state-of-the-art approaches on the KITTI dataset using 1) noising point cloud data  and 2) original point cloud data. For each task, three categories are involved (\textit{Cars}, \textit{Pedestrians} and \textit{Cyclists}). The comprehensive experimental results are reported for the three categories under the three difficulty levels (\textit{Easy}, \textit{Moderate} and \textit{Hard}).

\textbf{Results on noising point cloud data.} Extra challenge is  introduced for detection by adding some noise points to each object. We think it is a relatively reasonable to sample these noise points that are closer to real scenes, since farther noising points will not interfere with the detection of a certain object. Specifically,
the $x$, $y$ and $z$ coordinates of these noise points obey the uniform distribution of $[-l/2+c_x,-3l+c_x] \cup [l/2+c_x,3l+c_x]$, $[-h/2+c_y,-3h+c_y] \cup [h/2+c_y,3h+c_y]$ and $[-w/2+c_z,-3w+c_z] \cup [w/2+c_z,3w+c_z]$, respectively.
All models are trained with the official training data while tested on the noising validation point cloud data to evaluate their robustness for noises. 

Quantitative results with state-of-the-art methods  are presented in Table.~\ref{Noise_Experiments}. Although PointRCNN~\cite{shi2019pointrcnn} outperforms our method by 0.43\% in terms of 3D mAP for Cars, our method shows its superior robustness for noises. With 100 noise points added, our method yields a 3D mAP of 79.34\%, outperforming PointRCNN by 1.7\%. For Pedestrians, our method achieves an improvement of 5.8\% and 11.9\% comparing with PointPillars~\cite{lang2019pointpillars} and PointRCNN. It can be observed that our method demonstrates great robustness for noises,  especially for hard examples, \textit{e.g.,} Pedestrians, hard Cyclists and hard Cars.

On the whole, voxel-based methods~(\textit{e.g.,} PointPillars and our framework) are more robust to noising points compared with the method based on the raw point clouds~(\textit{e.g., PointRCNN}). The main reason is that PointRCNN is a two-stage method that first optimizes the Region Proposal Network~(RPN) separately and then optimizes the refinement network (\textit{i.e.,} RCNN) while fixing the parameters of RPN. For the contrast, our method is a coarse-to-fine detection network that can be end-to-end trainable, which is more robust for interference feature. 

\textbf{Results on original point cloud data.} 
The experimental results on the official KITTI test dataset are presented in Table.~\ref{3D_Object_Detection}. Our method yields an 3D mAP of 62.00\% over the three categories, outperforming the state-of-the-art methods PointPillars~\cite{lang2019pointpillars} and PointRCNN~\cite{shi2019pointrcnn} about 1.20\% and 1.22\%, respectively. In particular, for challenging objects (e.g. Pedestrians), our method achieves an improvement of 2.30\% and 4.83\% over PointPillars and PointRCNN, respectively. In addition, the visualization on the test set is provided in the supplemental material.

\textbf{The visualization of TANet.} We present the visualization of learned feature maps and predicted confidence scores feature produced by PointPillars~\cite{lang2019pointpillars} and TANet in Fig.~\ref{Vis_Attention}. It can be observed from the visualized feature maps that TANet can focus on salient objects and ignore noising parts. Besides, compared to PointPillars, our method outputs higher confidence scores on the salient objects. For more challenging objects which PointPillars fails to detect, our method still obtains a satisfactory confidence score. This explains why our method performs better than PointPillars.

\textbf{Running time.} The average inference time of our method is 34.75ms, including 1) 8.0 ms for data pre-processing; 2) 12.42ms for voxel feature extraction; and 3) 14.33ms for detection and the post-processing procedure.

\subsection{Ablation Studies}

In this section, extensive ablation experiments are provided  to analyze the effect of different components of TANet. All of our experimental results are conducted on the official validation set with 100 noise points for all the three categories. The official code of PointPillars~\cite{lang2019pointpillars} is reproduced as our baseline. Moreover, some experimental results can be seen on the official validation set without noise points on supplemental material.

\textbf{Analysis of the attention mechanisms.} Table.~\ref{Different_Feature} presents extensive ablation studies of the proposed attention mechanisms.  We remove the TA and Fine Regression(FR) module from our model as the baseline and achieves a 3D mAP of 65.59\%. With only Point-wise Attention~(PA) and only Channel-wise Attention~(CA), the performance is boosted to 67.04\% and 66.93\%, respectively.
And we name the parallel attention fusion mechanism for PA and CA as PACA. As shown in Table.~\ref{Different_Feature}, specifically, when combining them together, PACA yields a 3D mAP of 67.38\%, outperforming the baseline model by 1.8\%. 
To verify the superiority of PACA better, we provide three alternatives (\textit{Concat}, \textit{PA-CA} and \textit{CA-PA}), which combine the spatial attention and channel-wise attention in different ways. Concretely, \textit{Concat} concatenates the outputs of these two kinds of attention along the channel-wise dimension. PA-CA (resp. CA-PA) represents cascading spatial attention (resp. channel-wise attention) and channel-wise attention (resp. spatial attention) sequentially. It can be obviously observed that PACA outperforms all these three combination mechanisms, suggesting PACA is of great importance for making rational use of both spatial and channel-wise information. It should be noted that all these different ways for attention combination bring significant improvements over the baseline. Moreover, based on the PACA, TA module takes the voxel-wise attention into consideration, which achieves the improvement with about 0.40\%. It demonstrates the effectiveness and robustness of unifying the channel-wise attention point-wise attention and voxel-wise attention.

\begin{table}[h]
\scriptsize
\centering
\begin{tabular}{l|c|c|c|c}
\hline
 Method &Cars &Pedestrians &Cyclists &3D mAP\\
\hline
	Baseline &77.20 &55.04  &64.53 &65.59 \\
	CA &77.46 &57.61 &65.71  &66.93                 \\    
	PA &77.58 &57.59 &65.97  &67.04                 \\  
    PA-CA &77.90 & 56.65 &66.10 &66.88  \\
    CA-SA & 77.82&56.92&66.04 & 66.93  \\
    Concat &77.87 & 57.32 & 66.17 & 67.12 \\
\hline
    PACA~(Ours) &77.97 &57.94 &66.22  & 67.38 \\
    TA module~(Ours) &\textbf{78.33} & \textbf{58.43} & \textbf{66.61} & \textbf{67.79}\\
\hline
\end{tabular}
\caption{Ablation experiments on the effect of channel-wise attention, point-wise attention and voxel-wise attention, as well as different combination settings. All the experiments are conducted without FR module.}
\label{Different_Feature}
\end{table}

\textbf{Effect of the PSA module.} 
The effect of the PSA module is further explored. Our method also compares with the most representative single-shot refinement network RefineDet~\cite{zhang2018single} under two settings: with and without the TA module. Under the same setting, our method is 
consistently better than RefineDet. Without the TA module, it is noticed that the improvement of  the PSA module is not so obvious. But surprisingly, with the TA module, the performance of the  PSA achieves an evident improvement. It means that the PSA module has a fine complementarity with the TA module. The TA module can provide a robust and discriminate feature representations, and the PSA module can make full use of them to estimate 3D bounding boxes.

\begin{table}[h]
\scriptsize
\centering
\begin{tabular}{l|c|c|c|c}
\hline
 Method &Cars &Pedestrians &Cyclists &3D mAP\\
\hline
	Baseline &77.20 &55.04  &64.53 &65.59 \\
	Baseline + RefineDet~\shortcite{zhang2018single} &77.27 &55.86 &65.23 &66.12 \\
	Baseline + PSA &77.30 &56.02 &65.30 &66.21 \\\hline
    TA module &78.33 &58.43 & 66.61 &67.79\\
    TA module + RefineDet~\shortcite{zhang2018single} &79.28 &60.05 & 67.06 &68.80 \\
    TA module + PSA &79.34 &60.84  &67.87 &69.35\\
\hline
\end{tabular}
\caption{Ablation experiments on the effect of the proposed PSA module.}
\label{T3}
\end{table}

\section{Conclusion}
This paper proposes a novel TANet for 3D object detection in point clouds, 
especially for noising point clouds. 
The Triple Attention (TA) module and the Coarse-to-Fine Regression (CFR) module are the core parts of TANet. 
The former adaptively enhances crucial information of the objects and suppresses the interference points. 
The latter provides more accurate detection boxes without excessive computation cost. 
Our method achieves state-of-the-art performance on the KITTI dataset. 
More importantly, in the more challenging cases that the additional noise points are added, extensive experiments further demonstrate the superior robustness of our method, which outperforms existing approaches by a large margin.


\bibliography{egbib}
\bibliographystyle{aaai}


\clearpage
\section{Supplemental Material}

\subsection{The Visualization on Test Set}
\begin{figure}[h]
\begin{center}
  \includegraphics[width=1.0 \linewidth]{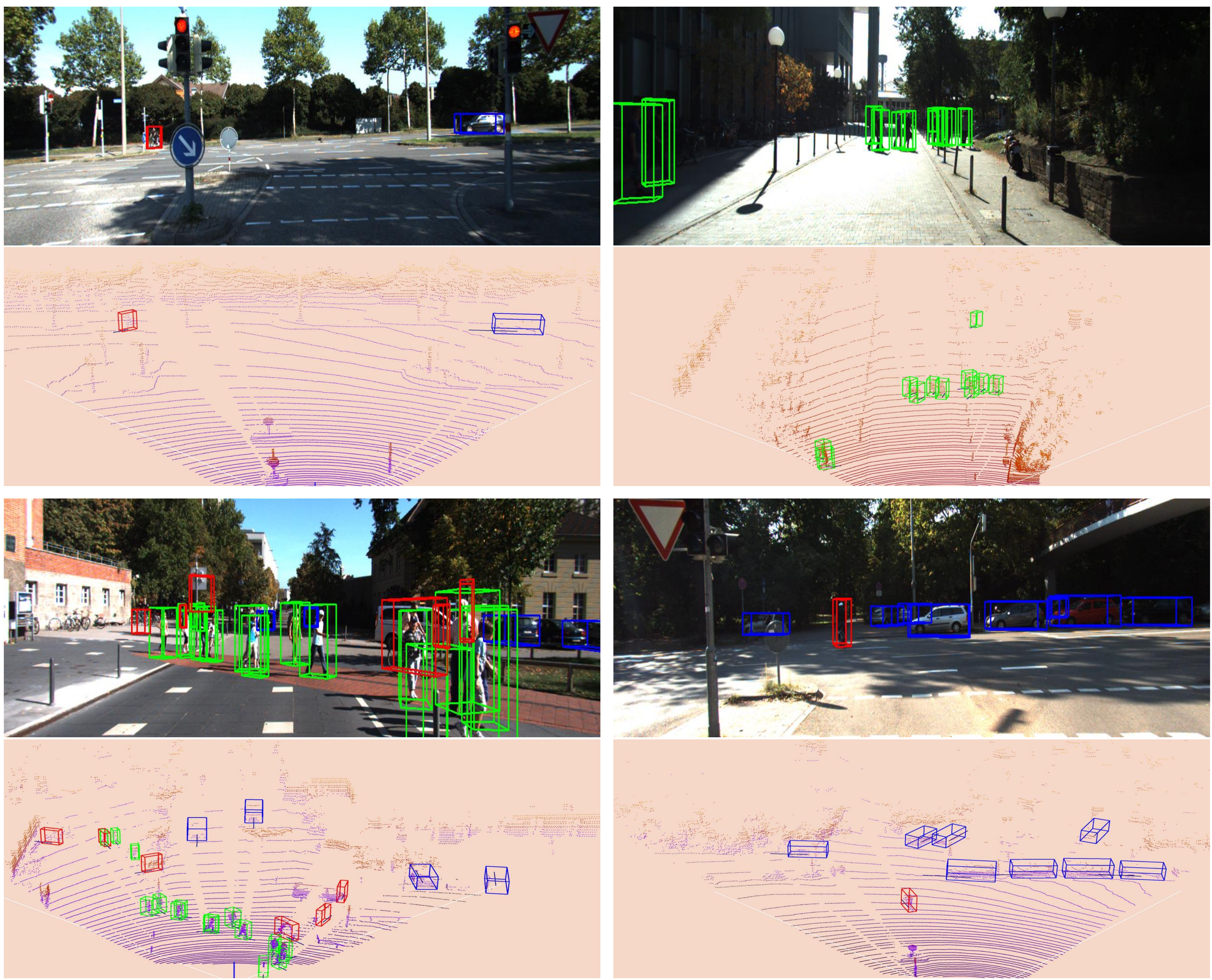}
\end{center}
\caption{Qualitative detection results on the KITTI test set. The \textbf{first} and the \textbf{third} row show the 3D bounding boxes projected on the 2D images. The \textbf{second} and the \textbf{fourth} row depict the predicted 3D bounding boxes and corresponding orientations for 3D objects on LiDAR. \textit{Cars}, \textit{Pedestrians} and \textit{Cyclists} are visualized with blue, green and red bounding boxes, respectively.}
\label{Vis_Results}
\end{figure}

Several qualitative results for 3D detection are shown in Fig.~\ref{Vis_Results}. It can be observed that our network can accurately predict both the locations and orientations of 3D objects even under extremely challenging situations (e.g., small objects and objects with heavy occlusions).

\subsection{More Implementation Details}
For the \textit{Cars} detection task, we consider the point clouds in the range of $[0,70.4]\times[-40,40]\times[-3,1]$ meters along the X, Y, Z axis, respectively. The voxel size is set to $v_W=0.16$, $v_H=0.16$, $v_D=4$ meters. Thus, the point cloud space is partitioned into a $440 \times 500 \times 1$ voxel grid. For each voxel, we adopt a single anchor box with two orientations (0 and $\pi/2$ radians). Concretely, we set $w_a=1.6$, $l_a=3.9$, $h_a=1.56$. We regard an anchor as positive if it achieves the highest IoU score with a ground truth box or an IoU score higher than 0.6. An anchor is considered as negative if its IoU score with each ground truth box is less than 0.45. We do not care about the anchors if their IoU scores are in the range of [0.45, 0.6] with all the ground truth boxes. Specifically, anchors containing no points are simply ignored for efficiency. In the post-progress step, we select the NMS score of value 0.3 and the IoU threshold of value 0.5, respectively. 

For \textit{Pedestrians} and \textit{Cyclists} detection tasks, the range of input point cloud is of $[0,48]\times[-20,20]\times[-2.5,0.5]$ along the X, Y, Z axis, respectively. We adopt the voxel size of $0.16 \times 0.16 \times 3$, yielding a voxel grid of size $300 \times 250 \times 1$. The anchor size is set to $w_a=0.6, l_a=1.76, h_a=1.73$. An anchor is considered as positive if it achieves the highest IoU score with a ground truth box or an IoU score higher than 0.5. We consider an anchor as negative if its IoU score with each ground truth box is less than 0.35. We do not care about the anchors if their IoU scores are in the range of [0.35, 0.5] with all the ground truth boxes. And the NMS score and the IoU threshold are set to 0.1 and 0.6, respectively.

\subsection{More Experiments}
In this part, we also provide some experimental results on the official validation set without noise points of the KITTI dataset.

\textbf{Results on Official Validation Set.} In addition, we report the results on the official validation set without noise points of the KITTI dataset for the convenience of comparisons with more future works in Table.~\ref{val_results}. 

\begin{table}[h]
\scriptsize
\centering
\begin{tabular}{l|c|c|c|c}
\hline
 Benchmark & Easy & Moderate & Hard & mAP\\
\hline
	Cars~(3D Detection) &88.21 &77.85 &75.62  &80.56\\
	Cars~(BEV Detection) &90.17 &87.55 &87.14 &88.29\\
    Pedestrians~(3D Detection)  &70.80  &63.45  &58.22 &64.16\\
    Pedestrians~(BEV Detection)  &76.70 &70.76 &65.13 &70.86\\
    Cyclists~(3D Detection)    &85.98&64.95&60.40 &70.44\\
    Cyclists~(BEV Detection)  &87.17& 66.71 &63.79  &72.56\\
\hline
\end{tabular}
\caption{The performance on the official KITTI validation set for \textit{Cars}, \textit{Pedestrians} and \textit{Cyclists}.}
\label{val_results}
\end{table}

\textbf{Selection of $\mathbf{\lambda}$.} We conduct extensive experiments by varying the balance weight $\mathbf{\lambda}$ in the validation set. In Table.~\ref{lambda_influence}, it can  be seen that $\lambda=2.0$ is the best choice and achieves 71.72\% 3D mAP.
\begin{table}[h]
\scriptsize
\centering
\begin{tabular}{l|c|c|c|c}
\hline
 $\lambda$ & 1.0 &2.0 & 5.0 & 10.0\\
\hline
 3D mAP &70.69 &71.72  &71.44 &70.80 \\
\hline
\end{tabular}
\caption{Analysis of the influence of the balance weight $\lambda$ for the multi-task loss.}
\label{lambda_influence}
\end{table}

\end{document}